\documentclass[10pt,letterpaper]{article}

\usepackage{cogsci}

\cogscifinalcopy % Uncomment this line for the final submission 

\usepackage{times}
\usepackage{apacite}
\usepackage{float} % Roger Levy added this and changed figure/table
\usepackage{graphicx}
\usepackage{amsmath}
\usepackage{amssymb}
\usepackage{booktabs}
\usepackage[dvipsnames]{xcolor}
\usepackage{lipsum}
\usepackage{multicol}
\usepackage{multirow}
\usepackage{epigraph}
\usepackage{natbib}

\usepackage{mathbbol}
\usepackage[OT1]{fontenc}
\usepackage[hyphens]{url}
\usepackage[colorlinks = true,
           linkcolor = CadetBlue,
           urlcolor  = BlueViolet,
           citecolor = BrickRed,
           anchorcolor = blue]{hyperref}
\usepackage{misra}
\definecolor{yello}{HTML}{ffb677}
\definecolor{blu}{HTML}{005082}
\definecolor{purpl}{HTML}{726a95}
\definecolor{orang}{HTML}{ff9a76}
\definecolor{tealish}{HTML}{1aa6b7}

\newcommand{\adaptation}{\mathcal{A}}
\newcommand{\generalization}{\mathcal{G}}
\newcommand{\concepts}{\mathcal{C}}
\newcommand{\properties}{\mathcal{P}}
\newcommand{\true}{\mathsf{True}}
\newcommand{\false}{\mathsf{False}}
\newcommand{\metric}{\mathrm{G}}

\newcounter{argument}
\Roman{argument}
\newenvironment{argument}[2]{\refstepcounter{argument}\equation\begin{tabular}{@{}l@{}}
        #1 \\ \midrule #2
    \end{tabular}}{\tag{\roman{argument}}\endequation}

\makeatletter
\renewcommand{\paragraph}{%
  \@startsection{paragraph}{4}%
  {\z@}{1ex \@plus 1ex \@minus .2ex}{-1em}%
  {\normalfont\normalsize\bfseries}%
}
\makeatother

\title{A Property Induction Framework for Neural Language Models}
\author{
{\large \bf Kanishka Misra,$^\textbf{1}$ Julia Taylor Rayz,$^\textbf{1}$ and Allyson Ettinger$^\textbf{2}$}\\
  $^1$Department of Computer and Information Technology,
  Purdue University, IN, USA \\
  $^2$Department of Linguistics, University of Chicago, IL, USA\\
  \texttt{kmisra@purdue.edu, jtaylor1@purdue.edu, aettinger@uchicago.edu}
}

\begin{document}

\maketitle

\begin{abstract}
To what extent can experience from language contribute to our conceptual knowledge? 
Computational explorations of this question have shed light on the ability of powerful neural language models (LMs)---informed solely through text input---to encode and elicit information about concepts and properties. To extend this line of research, we present a framework that uses neural-network language models (LMs) to perform property induction---a task in which humans generalize novel property knowledge (\textit{has sesamoid bones}) from one or more concepts (\textit{robins}) to others (\textit{sparrows, canaries}). Patterns of property induction observed in humans have shed considerable light on the nature and organization of human conceptual knowledge. Inspired by this insight, we use our framework to explore the property inductions of LMs, and find that they show an inductive preference to generalize novel properties on the basis of category membership, suggesting the presence of a taxonomic bias in their representations.

\end{abstract}

\section{Introduction}

There has recently been a growing interest in exploring the limits and potential of language as an environment for learning conceptual knowledge \citep{elman2004alternative, lupyan2019words}---knowledge that encompasses mental representations of everyday objects/events, and their properties and relations, that together inform our intuitive understanding of the world \citep{murphy2004big, machery2009doing}.
Computational explorations of this claim often study the extent to which models that learn semantic representations through text alone can capture conceptual knowledge \citep{lucy-gauthier-2017-distributional, forbes2019neural, da-kasai-2019-cracking, bhatia2020transformer}.

A hallmark feature of the conceptual knowledge acquired by humans is its capacity to facilitate inductive generalizations: inferences that go beyond available data to project novel information about concepts and properties \citep{osherson1990category, chater2011inductive, hayes2018inductive}.
For example, our knowledge of taxonomic specificity is reflected when we generalize a novel property of a concept (e.g., \textit{robins have T9 hormones}) more strongly to taxonomically close concepts (\textit{sparrows have T9 hormones}) than to more taxonomically distant concepts (\textit{tigers have T9 hormones}).
Inductive generalizations about novel properties (also called \textit{property induction}) therefore provide a context within which we can explore the nature of agents' understanding of conceptual knowledge.
In this paper, we develop an analysis framework that uses neural network-based language models (LMs, henceforth) to perform property induction and use this framework to study concept representation in these models.
Our framework consists of two stages. 
In the first stage, we train LMs to evaluate the truth of sentences expressing property knowledge (e.g., \textit{a cat has fur} $\rightarrow$ True, \textit{a table has fur} $\rightarrow$ False). 
In the second stage, we use these property-judgment models to test how the representations from the underlying LMs drive inductive generalization of novel properties---e.g., \textit{has feps, can dax,} etc.

Each stage of our framework sheds light on different aspects of the conceptual knowledge captured by LMs.
Using the first stage, we test the extent to which LMs support judgement of whether a property applies to a concept, even when that property has not been seen in task-specific fine-tuning. 
We find that LMs perform substantially above chance, consistent with the conclusion that they are able to rely on generalizable property knowledge to assess truth of concept-property associations.
In the second stage, we use this property judgment framework to study how knowledge representation in the base LMs drives inductive generalization with respect to entirely novel properties. 
We focus specifically on whether models' inductive preferences indicate reliance on taxonomic information, by testing whether models prefer to generalize within rather than outside of taxonomic categories. 
To do this, we teach our property-judgment models novel property information such as \textit{robins can dax} via standard backpropagation methods and then test the extent to which they prefer generalizing this novel property to other birds (e.g. \textit{sparrows can dax}) more strongly than to non-birds (e.g. \textit{zebras can dax}).
We find that models indeed show a preference for projecting new property knowledge on the basis of taxonomic category membership, suggesting that the models have acquired and represented taxonomic features on which they rely to project novel information.
% Additional experiments
% \todo{(effect of monotonicity) (effect of feature overlap observed during training) (taxonomic preference persisted even in cases where feature overlap was in conflict with category-membership, suggesting..).}

Our LM-based account of property induction contributes to the field in four primary ways. 
On the basis of the goals of the task, our framework focuses on reasoning where conclusions do not deductively follow from the premise, unlike the goals of the more commonly-used task of natural language inference \citep{bowman2015large}, and it therefore allows for testing of human-like inferences that are seldom studied in LMs \citep[cf.][]{bhagavatula2019abductive}.
Next, as we show below, our framework opens a new window into exploring how large neural network models of language generalize beyond their training experience, complementing inquiries of models' inductive bias with respect to syntactic structure \citep{mccoy2020does} and ``universal linguistic constraints'' \citep{mccoy2020universal}. Additionally, this work advances research aiming to diagnose the nature and extent of conceptual knowledge in LMs \citep{da-kasai-2019-cracking,forbes2019neural,weir2020probing, bhatia2020transformer} by additionally focusing on how knowledge present in LM representations drives the generalizations they make. 
Finally, at a high level, our framework contributes to a range of works that have applied connectionist models to the problem of property induction \citep[see][]{sloman1993feature, rogers2004semantic, saxe2019mathematical}.

\section{Testing Property Induction with Arguments}
Property induction is often studied in humans through the use of arguments, represented in the following premise-conclusion format, as popularized by \citet{osherson1990category}:
\begin{argument}
{Robins have sesamoid bones.}{All birds have sesamoid bones.}\label[arg]{arg:example}
\end{argument}
\Cref{arg:example} is read as \textit{``Robins have sesamoid bones. Therefore, all birds have sesamoid bones.''}
The subject of the premise sentence (\textit{robin}) is referred to as the premise concept (similarly, if there are multiple premises, we have a set of premise concepts), while that of the conclusion is called the conclusion concept.
Representing induction stimuli as arguments allows one to use the notion of ``argument strength,'' which quantifies the degree to which a human subject's belief in the premise statements strengthens their belief in the conclusion \citep{osherson1990category}.
In many cases, researchers control the type of novel properties provided to participants by using \textit{blank} properties---properties that are synthetically created and are therefore unknown to participants, maximizing the chances that they will use their knowledge of the relations between the premise and conclusion concepts to make generalizations \citep{rips1975inductive, osherson1990category, murphy2004big}. In our property induction experiments, we simulate blank properties by using nonce words to synthetically construct novel properties---e.g., \textit{can dax}, \textit{is vorpal}, etc and use them to explore knowledge of conceptual relations in LMs.

\section{The Framework}
Computationally, property induction can be viewed as making conditional probability estimates about the conclusion ($c$), given some premise ($\pi$): $p(c\mid\pi)$.
We interpret this measure in our framework as a probability that a novel property is applied to a conclusion concept, by a model whose representations reflect the premise information.
This interpretation leads to two desiderata that our framework aims to satisfy: (1) the ability to make judgments about the association of properties to concepts, and (2) the ability to accept new property knowledge and then be queried to assess generalization of this new property knowledge to additional concepts.
To satisfy (1), we fine-tune existing pre-trained LMs to classify as true or false sentences that associate properties to concepts---i.e., make property judgments. Doing so enables the LMs to estimate the probability that a property applies to a concept, as $p(\true \mid \textit{``concept has property''}, \phi)$, where $\phi$ stands for the parameters of a given LM. 
We use this approach rather than estimating sequence probabilities---which are relatively more straightforward to compute using LMs---in order to avoid surface-level confounds as observed in similar work by \citet{misra2021typicality}.
Next, to satisfy (2), we operationalize induction as the behavior of these LMs (now fine-tuned to make property-judgments) after further adaptation to new properties using standard backpropagation \citep{rumelhart1986learning}. 
The motivation to use backpropagation to perform property induction is simple---it allows the integration of new information in the model by directly updating its representations, which encode knowledge used to inform how the model generalizes.
Under this operationalization, we first adapt our LM to reflect the premise information (e.g. $\pi = \textit{a robin can dax}$) and then use the updated parameters of the model ($\phi^{\prime}$) to estimate the probability of a conclusion ($c = \textit{a sparrow can dax}$) as: $p(\true \mid c, \phi^{\prime})$.
% Furthermore, the updates that backpropagation introduces in the model can be directly quantified, allowing us to explore numerically how models integrate new information in their representations.
A similar operationalization of induction was used by \citet{rogers2004semantic}, who reported inductive inferences made by their PDP model of semantic cognition by updating its weights to reflect novel information, which was provided after several steps of training on general conceptual knowledge derived from a toy dataset  of concepts and properties.
{Similar methods have also been used by \citet{van2018neural} and \citet{kim-smolensky-2021-testing} to characterize the adaptation of grammatical knowledge in LMs.}
% Similar works by \citet{van2018neural} adapt language models on novel syntactic structures to shed light on their syntactic adaptation capacities.
\begin{figure*}[t!]
    \centering
    \includegraphics[width=0.9\textwidth]{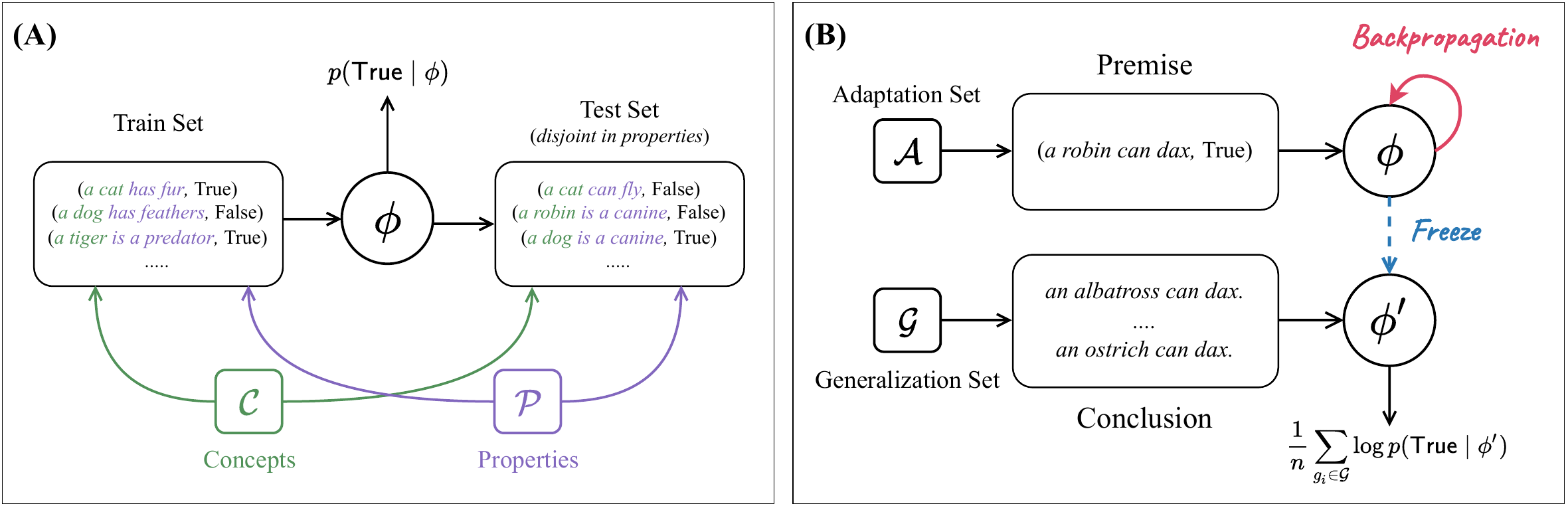}
    \caption{\textbf{(A)} Property Judgment Stage describing the training of the property judgment model (with parameters $\phi$) to make judgments of truth on sentences expressing concept-property assertions. Sentences created using the concept ($\concepts$) and property ($\properties$) data collected by \citet{devereux2014centre}; \textbf{(B)} Depiction of the Induction Stage, in this case, for testing the generalization of the novel property \textit{can dax} from robin to all birds. Here, $\adaptation = \{\textsc{robin}\}$, $\generalization = \{\textsc{albatross}, ..., \textsc{ostrich}\}$.}
    \label{fig:framework}
    \vspace{-1em}
\end{figure*}
We now explain the two stages of our property induction framework in greater detail:
\subsection{Stage 1: Eliciting Property Judgments using LMs}
In the first stage, we constrain LMs to explicitly rely on property knowledge by distinguishing correct (\textit{cat has whiskers}) and incorrect associations (\textit{sparrow has whiskers}) between properties and concepts. 
We do this by fine-tuning LMs to classify sentences that express concept-property associations to be true or false. Importantly, we fine-tune models in a way that keeps the evaluation sets disjoint in terms of properties---i.e., the model is trained to assess the properties \textit{has feathers, has a tail} and then tested on a distinct set of properties: \textit{can fly, has a beak}.
Therefore, in order to succeed on this task (i.e., minimize loss on a disjoint evaluation set), a model must rely on property knowledge encoded in its representations, to enable judgments about properties never seen during fine-tuning.
In experiments that follow, we verify the extent to which the models are indeed able to draw on generalized property knowledge in order to succeed in this task. 
Importantly, this stage assumes the presence of a repository of concepts ($\concepts$) and associated properties ($\properties$) as data for training and testing the model. 
We create sentences that express property knowledge by pairing properties from $\properties$ to concepts from $\concepts$. We then fine-tune the LM to classify these sentences as true or false. At the end of this stage, we have a trained model (with parameters $\phi$) that takes as input a sentence $s$ and produces a probability score $p(\true \mid s, \phi)$ corresponding to the degree of truth of $s$ as internalized by the LM.
\Cref{fig:framework}A illustrates the property judgment stage.

\subsection{Stage 2: Induction as Adaptation to New Knowledge}
In this stage (see \Cref{fig:framework}B), we use the fine-tuned model from the previous stage to perform property induction, which we operationalize as the behavior of the model after adaptation to new property knowledge via backpropagation.

A property induction trial involves (1) a set of premise concepts (which we denote as the adaptation set $\adaptation \subset \concepts$); (2) a set of conclusion concepts (denoted as the generalization set $\generalization \subset \concepts$); and (3) a novel property being generalized from the premise to the conclusion.
% The content of these artifacts depends on the specific inductive phenomena being simulated.
We construct sentences that associate the novel property to the concepts in $\adaptation$ and $\generalization$, yielding the premise and conclusion stimuli, respectively (see Figure 1B).
To perform property induction, we first adapt the model's parameters $\phi$ to the premise sentences by using standard backpropagation, yielding an updated state of the model, $\phi^{\prime}$, that correctly attributes the concepts in $\adaptation$ with the novel property.
We then freeze $\phi^{\prime}$ and query the model with the conclusion sentences to obtain the (log) probability of generalizing (or ``projecting'') the novel property to the concepts in $\generalization$. 
We refer to this measure as the ``generalization score'' ($\metric$)---i.e., the strength of projecting the novel property to a set of one or more concepts in the generalization set:
\begin{align}
    \metric = \frac{1}{n}\sum_{\textit{c}_i \in \mathcal{G}} \log p(\true \mid \textit{``c}_i\textit{ has property X''}, \phi^{\prime})\label{eq:genscore}
\end{align}
The model parameters are reset to their original state ($\phi$) after this step in order to perform subsequent trials. 

We now use components of this framework in two experiments---one for each stage in the framework.

\section{Investigating LMs on Property Judgments}
Our first experiment focuses on the first stage of the proposed induction framework. Here, we fine-tune pre-trained LMs to evaluate the truth of sentences attributing properties to concepts---i.e., we want our models to map the sentence \textit{a cat has fur} to $\true$ and \textit{a cat can fly} to $\false$. We use an existing semantic property norm dataset to construct our sentences and split them into disjoint evaluation sets, where the properties we test the model on are strictly different from those the model sees during fine-tuning. Therefore, a model must learn to rely on its `prior' (pre-trained) property knowledge in combination with task specific information it picks up during fine-tuning in order to succeed on this task.

\paragraph{Ground-truth Property Knowledge Data}
To construct sentences that express property knowledge, we rely on a property-norm dataset collected by the Cambridge Centre for Speech, Language, and the Brain \citep[CSLB;][]{devereux2014centre}.
The CSLB dataset was collected by asking 123 human participants to elicit properties for a set of 638 concepts, and this dataset has been used in several studies focused on investigating conceptual knowledge in word representations learned by computational models of text \citep[e.g., ][]{lucy-gauthier-2017-distributional, da-kasai-2019-cracking, bhatia2020transformer}.
Importantly, property-norm datasets such as CSLB only consist of properties that are applicable for a given concept and do not contain negative property-concept associations.
As a result, the works that have used these datasets
% \citep{lucy-gauthier-2017-distributional, da-kasai-2019-cracking, bhatia2020transformer} 
sample concepts for which a particular property was not elicited and take them as negative instances for that property (e.g., \textsc{table, chair, shirt} are negative instances for the property \textit{can breathe}), which can then be used in a standard machine-learning setting to evaluate a given representation-learning model.

Upon careful inspection of the CSLB dataset, we found that the above practice may unintentionally introduce incorrect or inconsistent data.
Datasets such as CSLB are collected through human elicitation of properties for a given concept, so it is possible for inconsistencies to arise. One way that this may happen is if some participants choose not to include properties that are obvious for the presented concept (e.g., \textit{breathing} in case of living organisms), while other participants do, resulting in an imbalance that can be left unaccounted for.
We found that this was indeed the case: e.g., the property \textit{has a mouth} was only elicited for 6 animal concepts (out of 152), so all other animals in the dataset would have been added to the negative search space for this property during sampling, thereby propagating incorrect and incomplete data. This indicates a potential pitfall of directly using property-norm datasets to investigate semantic representations---and suggests that prior evaluations and analyses \citep{lucy-gauthier-2017-distributional, da-kasai-2019-cracking, bhatia2020transformer} may have falsely rewarded or penalized models in such cases. 
Owing to space constraints, we provide our detailed method and protocol to mitigate this problem in the supplemental materials. The revised dataset that we produce consists of a set of 521 concepts, corresponding to 23 different taxonomic categories (as annotated by the original authors of the CSLB dataset) and 3{,}735 properties, with 23{,}107 ground-truth property-concept pairs which we used in our experiment.

{For each of our 3{,}735 properties---associated with $k$ different concepts---we sample $k$ additional concepts that are maximally similar to the $k$ concepts associated with that property, and take these to be negative samples. For instance, for the concept \textsc{zebra}, we want to use \textsc{horse} for a negative sample rather than a more distant concept such as \textsc{table}.
By doing this, we make the property judgment tasks more difficult, increasing the chances that the models we obtain from this stage focus on finer-grained conceptual/property knowledge as opposed to coarser-grained lexical similarity.
% indeed focusing on conceptual knowledge to make property judgments instead of relying on simpler superficial cues such as lexical co-occurrence \ake{might be more accurate to say this helps ensure the models focus on something like finer-grained conceptual/property knowledge rather than coarse-grained semantic similarity/association}. 
For selecting similar concepts we take the \textit{Wu-Palmer similarity} as our similarity function \citep{wu-palmer-1994-verb}, which we compute over the subset of the WordNet taxonomy \citep{miller1995wordnet} that contains the senses of the 521 concepts considered in our experiments. We then follow the method outlined by \citet{bhatia2020transformer} to convert our 46{,}214 property-concept pairs (23{,}107 $\times$ 2) into natural language sentences, which we then use as inputs to our models. We split these sentences (paired with their respective labels) into training, validation, and testing sets (80/10/10 split), such that the testing and validation sets are only composed of properties that have never been encountered during training (note that properties between training and validation sets are also disjoint). 
We do this to avoid data leaks, and to ensure that we evaluate models on their capacity to learn property judgment as opposed to memorization of the particular words and properties in the training set. We make our negative sample generation algorithm and the resulting dataset of property-knowledge sentences available in our supplementary materials.}

\paragraph{Tested LMs}
While our framework can be applied to any neural language model, we present results from fine-tuning three pre-trained LM families, based on the precedent of using these models in standard sentence classification tasks \citep{wang2018glue}: BERT \citep{devlin-etal-2019-bert}, RoBERTa \citep{liu2019roberta}, and ALBERT \citep{lan2019albert}. 
All three models use the transformer architecture \citep{vaswani2017attention}, and are trained to perform masked language modeling: the task of predicting masked words in context in a cloze-task setup, where models have access to context words to the left and right of the masked word. 
We report results using the largest models in each of the three families---BERT-large, RoBERTa-large, and ALBERT-xxl---since these variants had the best performance in our preliminary experiments (on a separate validation set).
We fine-tune each of the three models on the property knowledge data by minimizing their binary cross-entropy loss on the training set using the AdamW optimizer \citep{loshchilov2018decoupled}.
We tune the hyper-parameters of the LMs on the validation set, and evaluate the three adapted models on the test set using F1 scores.

\paragraph{Results}
\Cref{tab:f1} shows the performance of the three models in our property judgment experiments. 
% \km{generally high scores, suggesting that to an extent the model learn to rely on concept knowledge through this true-false setup.}
We find that all three models show similarly high performance on the test set (0.78-0.79), suggesting strong capacities of all three models to assess the application of properties to concepts. Notably, the ALBERT-xxl model shows the same performance as BERT-large and RoBERTa-large despite having $\approx$130M fewer parameters, suggesting that this property knowledge can be encoded in smaller models with more efficient use of parameters.
Furthermore, all three models perform significantly above chance ($p < .001$, FDR corrected).
% We additionally compare our models' performance to a simple heuristic-driven baseline, which predicts labels for a property based on the concept-label statistics in the training set. That is, if the word \textit{robin} is seen more frequently in training sentences with the $\true$ label as compared to $\false$, then this model predicts $\true$ every time it sees \textit{robin} in the test set, we refer to this as the ``Concept-bias" model. We find the performance of all three models to be greater than the concept-bias model using bootstrap hypothesis testing \todo{(p; FDR corrected)}.

\begin{table}[t!]
\vspace{-1em}
\def\arraystretch{1.15}
\centering
\caption{Performance (F1 score) of the fine-tuned LMs on the test set of the property judgment task. Chance F1 is 0.66.}
\label{tab:f1}
\vspace{0.5em}
\begin{tabular}{|l|c|c|c|}
\hline
\textbf{Model} & \textbf{Params} & \textbf{Test F1} \\ \hline
ALBERT-xxl     & 206M            & 0.79    \\
BERT-large     & 345M            & 0.78    \\
RoBERTa-large  & 355M            & 0.79    \\ \hline
\end{tabular}
\vspace{-1em}
\end{table}

\section{Investigating Taxonomic Generalizations in LMs using Property Induction}
Taxonomic relations between concepts have an important role in studies of human inductive reasoning.
Early evidence from \citet{gelman1986categories} indicated a strong preference of children and adults, when making generalizations about new and unfamiliar properties, to do so based on the structure of biological taxonomies and category membership.
Building on this, \citet{osherson1990category} documented 13 separate taxonomic phenomena that influenced inductions made by humans.
Inspired by these works, we demonstrate how our property induction framework can be used to test whether a similar taxonomic bias is reflected in the LMs used to train the above property judgment models. 
For instance, if a model is provided with a new property---e.g., \textit{can fep}---that is associated with the concept \textsc{cat}, to what extent do its representational biases cause it to prefer generalizing or projecting this property to other mammals rather than to fish? 

\paragraph{Data} We restrict our analysis to the animal-kingdom subset of the concepts in our modified property-norm data, corresponding to a total of 152 animal concepts.
We first select the top six categories within this subset: \textsc{mammal} (52), \textsc{bird} (36), \textsc{insect} (18), \textsc{fish} (14), \textsc{mollusk} (8), and \textsc{reptile} (7).
Each instance in this experiment involves one of the six aforementioned categories (of size $m$) from which we sample $n$ concepts to create the adaptation set, and use the remaining $m-n$ to create the ``Within-category'' generalization set.
We then create two separate ``Outside-category'' generalization sets.
First, we sample the top $m-n$ animal concepts, on the basis of their average cosine similarity with the concepts in the adaptation set (using the representations of the embedding layer of the given model), and take this to be the ``Outside$_{\textit{similar}}$'' generalization set.
We use this similarity-based sampling technique in order to increase our confidence that observed differences can be attributed to category differences and not to general co-occurrence properties as learned by the models. This choice makes the Outside$_{\textit{similar}}$ set model-dependent.
We then complement this with an equal sized ``Outside$_{\textit{random}}$'' generalization set which is model-independent and is composed of concepts randomly selected from the set of animal concepts (excluding those that belong to the main category used for adaptation).
% Similarly, we create the ``Outside-category'' generalization set by sampling the top $m-n$ animal concepts that are outside the focus category, on the basis of their average cosine similarity with the concepts in the adaptation set (calculated using the representations from the embedding layer of the given model). We use this weighted sampling technique in lieu of simple random sampling in order to increase our confidence that observed differences can be attributed to category differences and not to general similarity properties.
We repeat this sampling process 10 times for each of $n = 1, \dots, 5$ adaptation concepts, and 8 novel properties: verb phrases created using nonce words (\textit{can dax, can fep, has blickets, has feps, is a tove, is a wug, is mimsy, is vorpal}). In total, we have 2{,}400 adaptation trials per model, each involving 3 generalization sets: Within, Outside$_{\textit{similar}}$, and Outside$_{\textit{random}}$ to test the property induction behavior of our models. 

\begin{figure}[t!]
    \centering
    \includegraphics[width=0.9\columnwidth]{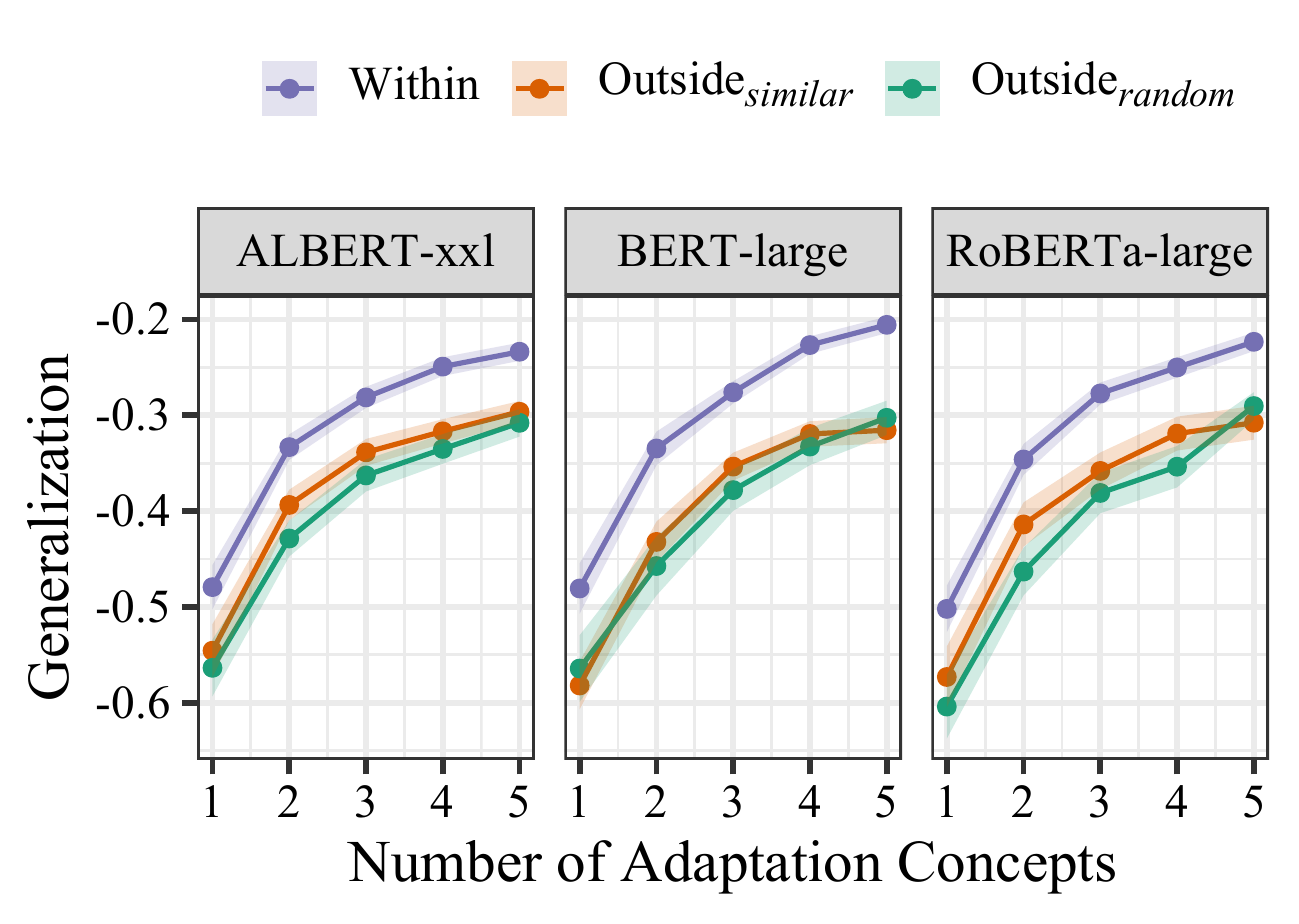}
    \vspace{-1em}
    \caption{Results from the taxonomic generalization experiment showing generalization scores ($\metric$) of the three property-judgment models for `Within' and both the `Outside' generalization sets across different number of adaptation concepts.}
    \label{fig:taxonomicresults}
    \vspace{-1.5em}
\end{figure}

\paragraph{Method}
In each trial, we pass the adaptation set to the models and let them minimize their loss (starting from the same optimizer state obtained at the end of the property judgment training phase) until they reach perfect accuracy. Then we compute $\metric$ for each of our three generalization sets as shown in \cref{eq:genscore}, for each model. 
\Cref{fig:taxonomicresults} shows the average $\metric$ (over all properties) as a function of the size of the adaptation set.

\paragraph{Results and Analysis}
We expect models with a preference for category-based generalization to have greater average $\metric$ value for the ``Within'' set than for either of the ``Outside'' sets.
From \Cref{fig:taxonomicresults}, we see that all three models consistently show this pattern---for all models, the average $\metric$ was significantly greater for ``Within'' generalization as compared to both ``Outside'' generalization sets ($p <.001$, according to a Games-Howell test conducted following a Welch's ANOVA). 
This suggests that these models show a preference for  generalizing newly-learned properties of a concept to other members of that concept's superordinate category.
% ALBERT-xxl and BERT-large consistently 
% show this pattern (differences between ``Within-category'' and ``Outside-category'' were significant at $p<.001$ using a t-test), suggesting that these models do show taxonomic biases in property generalization. This preference is substantially smaller for RoBERTa-large (with non-significant differences between generalization to within versus outside category for $n=1,2$), suggesting relative indifference in that model to taxonomic vs. representational similarity when it comes to extending new property knowledge to concepts outside the adaptation set.
We also observe that the average generalization score in both categories increases with an increase in the number of adaptation concepts. Notably, this is also robustly observed in humans (characterized as the \textit{premise monotonicity effect} by \citeauthor{osherson1990category})---however, we do not focus on this effect, as it is relatively expected that machine learning models will be more confident in their predictions as the number of samples provided to them increases.
% although it is interesting to note that this pattern is observed also for inputs that are not observed during adaptation.

Although the properties provided to the models in our induction experiment are ones that they have never seen during property-judgment training, one may wonder to what extent the models' inductive behavior can be explained based on taxonomically similar concepts simply being more likely to share properties within the property-judgment training stage---this could call into question how much these generalization patterns tell us about the underlying concept knowledge in the LMs.
Under the connectionist perspective of property induction \citep{sloman1993feature, rogers2004semantic}, the strength of generalization (of a novel property) to a concept is proportional to the overlap in properties between the premise (adaptation set) and the conclusion (generalization set).
We can reasonably expect this to translate to the models that we use here,
especially since they are trained to predict the presence and absence of properties.
To test the connection between LMs' generalization behavior and the overlap in training data properties, we first calculate property overlaps between each adaptation/generalization set pair as the ratio of the intersection and union of the ground-truth properties associated with the concepts within the sets (i.e., the jaccard similarity).
We then fit a linear mixed-effects model to predict $\metric$ using the \texttt{lme4} \citep{lme4} and \texttt{lmerTest} \citep{lmertest} packages in R, for each LM. 
Our final model included the number of adaptation concepts (\texttt{n}), as well as the property overlap (\texttt{overlap}) and cosine similarity (\texttt{sim}) between the adaptation and generalization sets along with their interaction as fixed effects; and also included random intercepts for the novel property and the trial. Model Specification: \texttt{G $\sim$ n + overlap * sim + (1|property) + (1|trial)}.
For all three LMs, we find a positive main effect of the property overlap,\footnote{along with that of number of concepts (\texttt{n}) as well as the model's cosine similarity (\texttt{sim}), $p < .001$ in all cases, approximated using Satterthwaite's method; see Suppl. Materials for full results.} suggesting that $\metric$ was significantly greater for generalization sets whose concepts had greater training data property overlap with those in the adaptation set.
% \todo{a conclusion?}

\begin{figure}[t!]
    \centering
    \includegraphics[width=0.8\columnwidth]{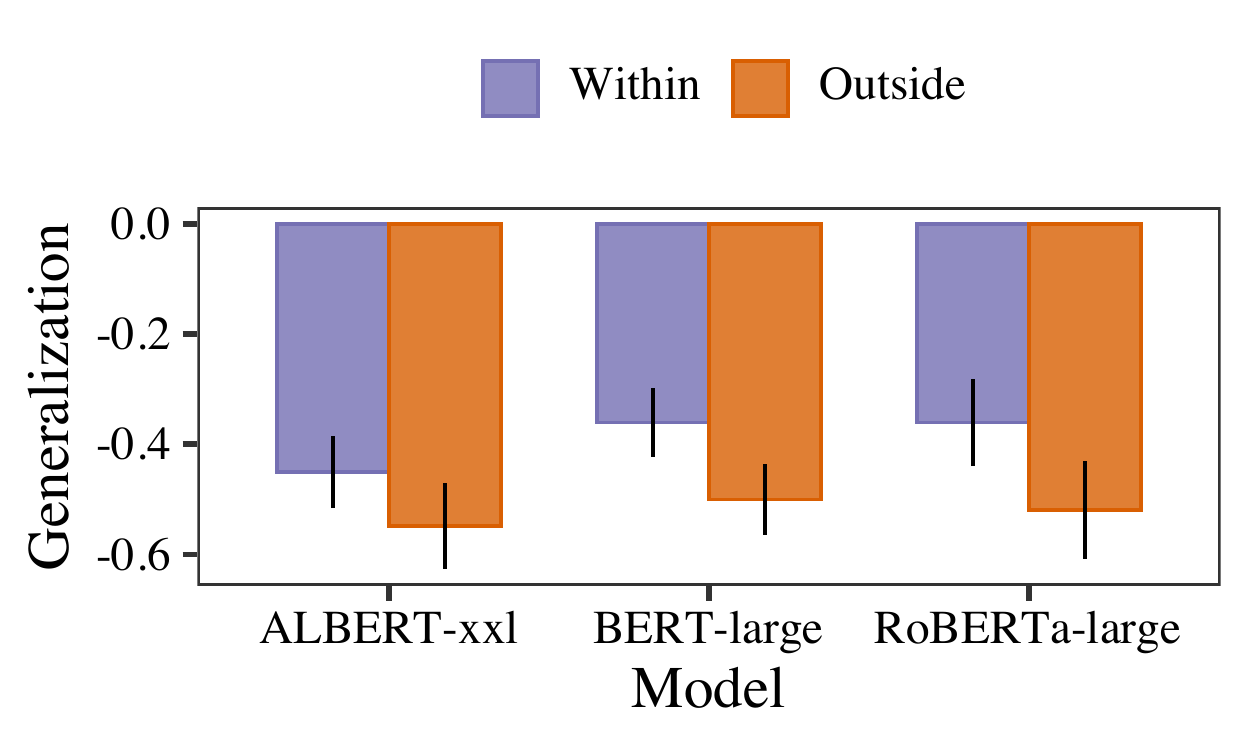}
    \vspace{-1em}
    \caption{Generalization scores of the models in cases where the Outside category had greater property overlap than the Within category. $N = 48$ trials for each model.}
    \label{fig:teaseapart}
    \vspace{-1.5em}
\end{figure}

While we have established that the models make generalizations that are consistent with the training set statistics, there exist cases where property overlap is in direct conflict with taxonomic category membership. For instance, dolphins share many salient properties with fish and yet are classified as mammals. 
Motivated by this observation, we ran another experiment involving only the cases where generalizations based on category membership conflicted with those based on property-overlap. We identified 6 concepts 
that had greater property overlap with concepts belonging to a different category relative to their own superordinate category: (\textsc{dolphin, whale, turtle, slug, snail, hippo}). 
For each of these 6 concepts, we compare generalization of our previously used 8 novel properties to concepts in the same taxonomic category (Within) vs. concepts in the category with which the concept had greater property overlap (Outside), thereby teasing apart the effect of property-overlap from that of true taxonomic membership.
\Cref{fig:teaseapart} shows results of this experiment. We observe for each model that the inductive preference for the ``Within'' generalization set was significantly greater than that for the ``Outside'' generalization set ($p < .001$ in all cases using a paired t-test, FDR corrected). 
This indicates that while the overall generalization behavior of the models is predicted by training data statistics, the models are robust in showing a taxonomic bias even when this relationship does not hold.

\section{General Discussion and Conclusion}
The empirical success of neural network-powered language models (LMs)---especially on high-level semantic tasks---has lent further support to the study of language as a source of semantic knowledge \citep{elman2004alternative, lupyan2019words}.
The goal of this paper was to contribute to this line of inquiry by understanding the ways in which  LMs generalize novel information about concepts and properties (\textit{a lion can fep}) beyond their training experience.
To this end, we developed a framework that used LMs to perform \textit{property induction}---a paradigm through which cognitive scientists have studied how humans use their conceptual repertoire to project novel information about concepts and properties in systematic ways \citep{rips1975inductive, osherson1990category, hayes2018inductive}.
By simulating a similar process in LMs, our framework can yield insights about the inductive preferences that are guided by the LMs' representations and shed light on the nature of the models' conceptual knowledge.

As a motivating case study, we used our property induction framework to study the extent to which LM representations show a preference to project novel properties on the basis of category membership. 
To this end, we adapted three LMs---fine-tuned to predict the truth of sentences expressing property knowledge---to inputs associating a novel property with one or more concepts. We then compared the models' projection of the novel property between (1) a set of concepts with the same superordinate category as the concept(s) associated with the property, and (2) a pair of concept-sets that were outside of that superordinate category.
In a majority of cases, the LMs preferred to project the new property to concepts of the same category, suggesting the influence of taxonomic bias. 
We hypothesized that some of models' taxonomic category preference could be due to high property overlap between concepts of the same category in property-judgment training---but while these property overlaps were statistically predictive of how models projected novel properties, the preference to generalize to concepts within the taxonomic category persisted even when effects of property-overlap and category-membership were teased apart.

Our results indicate that when LMs---fine-tuned to assess property knowledge---deploy knowledge about novel properties, they are guided in part by representational taxonomic biases beyond simple property-overlap relevant during fine-tuning. While we cannot say precisely what the source of this taxonomic bias is within these models, a simple explanation would be that this bias reflects the nature of the conceptual knowledge that these LMs learn and encode during pre-training. 
That is, in learning semantic representations of words by predicting them in context, models may have picked up on latent taxonomic knowledge, to which they then show sensitivity when projecting novel property information. 
This is consistent with existing works that diagnose  conceptual knowledge in LMs, finding them to display strong performance in predicting taxonomic category membership \citep{da-kasai-2019-cracking,bhatia2020transformer}. 
Through our results, we learn that this knowledge can additionally be implicitly activated, and in fact guides how new property information is generalized by LMs.

What other phenomena guide the inductive generalizations that LMs make about concepts and properties?
Our framework provides a flexible mechanism to simulate and test a broad range of phenomena observed in the human property-induction literature \citep[see][]{kemp2014taxonomy, hayes2018inductive}, and to shed light on the extent to which LMs' inductive preferences are consistent with those observed in humans. 
% \todo{intuitive theories guide property induction by providing the context on the basis of which humans reason. The taxonomic relationship is just one such intuitive theory out of many. Our framework can shed light on the extent to which LMs can represent the knowledge encompassed by these intuitive theories across various settings.}
A potential direction includes testing for a more general class of inductive phenomena that are guided by ``intuitive theories'' \citep{murphy1993theories} that provide the context on the basis of which different types of novel properties are projected differently \citep{carey1985conceptual, kemp2009structured}. For instance, biological information may be projected across a taxonomy (\textsc{robin} and \textsc{swan}), whereas behavioral information may be projected on the basis of specific shared properties (\textsc{hawk} and \textsc{tiger}).
Applying our framework on these phenomena can provide insight into the context-specific flexibility of LM representations, and at a high-level, the kinds of domain knowledge that can be acquired through text-exposure.

\paragraph{Acknowledgements} The authors thank the anonymous reviewers and the meta-reviewer for their feedback. K.M. is grateful to the CompLing lab (A.E.'s group) at the University of Chicago for discussion on previous iterations of the ideas presented here. The experiments reported in this paper were partially run on the Gilbreth cluster at Purdue University’s Rosen Center for Advanced Computing (two NVIDIA V100 GPUs with 32GB VRAM), and partially run on Hemanth Devarapalli’s computational platform consisting of a single NVIDIA 3090 GPU with 24GB VRAM.
% were run on  and to Hemanth Devarapalli for his assistance in accessing GPU compute (a single NVIDIA 3090 with 24GB of VRAM).

\paragraph{Reproducibility} We make our code and data available at \url{https://github.com/kanishkamisra/lm-induction}. This repository also consists of a supplementary materials document, \texttt{supplementary.pdf}, which contains details about our data filtering pipeline, the negative sampling generation algorithm and full set of results from our analyses.

\bibliographystyle{apacite}

\setlength{\bibleftmargin}{.125in}
\setlength{\bibindent}{-\bibleftmargin}

\bibliography{CogSci_Template}

\end{document}

\maketitle

% \begin{quote}
% \centering
%     Code and analyses: \url{https://github.com/kanishkamisra/lm-induction}
% \end{quote}
\section{Property Knowledge Re-annotation}
\paragraph{Premise}
Datasets such as the CSLB \citep{devereux2014centre} naturally lend themselves to investigations that probe the conceptual knowledge of computational models and their representations.
The CSLB dataset was collected by tasking 123 human participants to generate properties of a total of 638 concepts. For each property the authors then calculated its production frequency for all concepts for which it was generated, i.e., if the property \textit{can fly} was generated for the concept \textsc{robin} by 20 out of the 30 participants who were shown the concept, then its production frequency is 20. Note that the CSLB data set contains only positive property-concept associations. To construct negative samples, prior works that use CSLB as ground-truth to probe word representations typically use the set of concepts for which a given property was not generated, as negative \citep[e.g.][]{lucy-gauthier-2017-distributional, forbes2019neural, da-kasai-2019-cracking, bhatia2020transformer}. That is, negative samples are usually generated using concepts that have a production frequency of 0 for each property. Once a sufficient number of negative samples have been generated, the authors then train a probing classifier for every property, which predicts 1 if the production frequency of the property for that concept is nonzero, and 0 otherwise.

\paragraph{Limitation} Since the task that was employed to construct the CSLB dataset was that of generation as opposed to validation, it is possible---and perhaps likely---that it resulted in inconsistent annotations, where some humans might have forgotten to generate \textit{obvious} properties for certain concepts, or simply ignored them. For instance, the property \textit{can breathe}, which is obviously applicable for all animals, was missing in 146 animal concepts within the dataset. This means that if one were to follow the standard negative-sampling method described earlier, they would consider all 146 of these animals as concepts for which the property \textit{can breathe} does not hold true, which is incorrect. 
We conjecture that humans fail to generate features that are \textit{obviously valid} for certain concepts (e.g., \textit{can breathe, can grow, is a living thing} for animals) because they may be operating under Grice's maxim of quantity \citep{grice1989studies}, by only eliciting non-trivial or \textit{truly} informative properties for concepts in order to avoid redundancy.
While we leave the testing of the hypotheses within this conjecture for future work, this limitation of incomplete data raises questions about the extent to which we should trust the results and conclusions of prior work which are crucially affected by this problem, which we summarize using the aphorism: \textit{absence of evidence is not evidence of absence}.
% Since this is a generation task, humans miss out on obvious facts.. conjecture - Grice. Absence of evidence is not evidence of absence, but this problem persists in all works that use this dataset as ground-truth. 
% In what follows we describe our procedure to manually correct these inconsistencies.

\paragraph{Manual re-annotation of missing property-concept pairs}
% This is because they contain human-elicited properties of a number of concepts (638 to be precise). The typical task that humans were subjected to in the collection of CSLB was to generate property phrases for a fixed set of concepts. The authors then listed the `production frequency' or the number of 

To mitigate the limitation discussed above, we first selected the categories \citep[hand-annotated by][e.g., \textsc{bird, vehicle, tree}, etc.]{devereux2014centre} that had at least 9 concepts in the dataset and were not labeled as ``miscellaneous,'' resulting in 23 different categories with a total of 529 unique noun concepts, and 4{,}970 unique properties.
Next, we manually removed concepts and properties that contained proper nouns (e.g., \textsc{rolls-royce}, \textit{is in Harry Potter}), stereotypical or subjective data (e.g., \textit{is meant for girls}, \textit{is ugly}), and explicit mentions of similarity or relatedness (e.g., \textit{is similar to horse}). We further normalized properties that were paraphrases of each other (e.g., \textit{is used to flavor, does flavor} $\rightarrow$ \textit{is used to flavor}). This resulted in 521 concepts and  3{,}735 properties.
Again through manual search, we further identified a total of 365 properties that were incompletely annotated (i.e., those that were associated with certain concepts but were omitted for many relevant concepts during data collection---e.g., the property \textit{can grow} was missing for all invertebrates, despite being associated with all of them). 
We manually extended the coverage for these properties by adding in entries for concepts for which they had not been elicited. For instance, for the property \textit{can breathe}, which was generated for 6 out of 152 animals in the original dataset, we further add the remaining 146 concepts as additional positively associated concepts, increasing its coverage from 6 to 152.
While the total number of incompletely annotated properties is small (10\% of the valid properties), our re-annotation process greatly increases the total number of concept-property pairs (from 13{,}355 pairs in the original, unmodified dataset, to 23{,}107: an increase of 72\%) since many of the incompletely labeled properties were applicable across several categories (e.g., \textit{has a mouth, can grow,} etc).
After applying this process to the CSLB dataset, we are left with 23{,}107 property-concept pairs, which we use in subsequent experiments.
% The annotation process can be found in the file \texttt{re-annotation.R} in the github repository.\footnote{\url{https://github.com/kanishkamisra/lm-induction/R/re-annotation.R}} Furthermore, 
The re-annotated data can be found in the file \texttt{post\_annotation\_all.csv}\footnote{\url{https://github.com/kanishkamisra/lm-induction/data/post\_annotation\_all.csv}} in the github repository.

\paragraph{Final thoughts} The re-annotation process described above was performed manually due to resource, time, and financial constraints. However, we recommend running a large-scale empirical validation studies for datasets such as CSLB and McRae, before using them for probing experiments. 
While this is non-ideal in terms of resource use, it is necessary in order to draw faithful and appropriate conclusions about the correspondence between conceptual knowledge in humans and machines. 
Finally, a manuscript describing this process in greater detail, a small validation experiment ($\approx$2400 annotations) with humans, as well as empirical implications of the limitations described herein is in the works.

\section{Negative Sample generation using Taxonomies}
Here we describe our algorithm to generate negative samples for our first experiment in the paper---the property judgment task, where LMs are fine-tuned to classify as $\true$ or $\false$ sentences that attribute properties to concepts. For instance, the sentence \textit{a cat can fly} is labeled as $\false$ as \textsc{cat} is a negative sample for the property \textit{can fly}, whereas, \textit{a robin can fly} is labeled as $\true$.
Briefly, for the set of positive samples for a given property, we sample an equal-sized set of negative samples that are maximally similar to the positive samples. We use a taxonomic similarity (described below) as our similarity measure as it is model-free. Below we describe useful notation involved in the process, and then describe the full algorithm.
\subsection{Notation and Preliminaries}
\Cref{tab:notation} describes the notation we follow to construct our property judgment dataset. Our goal here is to generate 23{,}107 negative samples and then take the entire set of 46{,}214 concept-property pairs and their labels to carry out the property-judgment experiment.
% Please add the following required packages to your document preamble:
% \usepackage{booktabs}
% \usepackage{graphicx}
% Please add the following required packages to your document preamble:
% \usepackage{booktabs}
% \usepackage{graphicx}
\renewcommand{\arraystretch}{1.2}
\begin{table}[!ht]
\centering
% \resizebox{\textwidth}{!}{%
\begin{tabular}{@{}cp{10cm}p{3.5cm}@{}}
\toprule
\textbf{Notation} &
  \textbf{Meaning} &
  \textbf{Remarks} \\ \midrule
$\concepts$ &
  The set of all concepts in our experiments. These are also at the lowest level of the taxonomy---i.e., its leaf nodes. &
  $|\concepts| = 521$ \\
$\properties$ &
  The set of all unique properties used in our experiments. &
  $|\properties| = 3735$ \\
$\positives_{P_i}$ &
  The set of concepts that possess the property $P_i$. &
  $\positives_{P_i} \subset \concepts, |\positives_{P_i}| = k$ \\
$\leftovers_{P_i}$ &
  The set of concepts that do not possess the property $P_i$, i.e., $\leftovers_{P_i} = \concepts - \positives_{P_i}$ &
  $|\leftovers_{P_i}| = 521 - k$ \\
$\negsamp(\leftovers_{P_i}, k)$ &
  A function that extracts $k$ negative samples from $\leftovers_{P_i}$ using the method described in \Cref{alg:propjudgdataset} (lines 6--9). &
  $|\negsamp(\leftovers_{P_i}, k)| = k,$ \\ \bottomrule
\end{tabular}%
% }
\caption{Notation for various artifacts involved in the paper.}
\label{tab:notation}
\end{table}

In order to generate negative samples, we first tag the senses of all our 521 concepts using the WordNet \citep{miller1995wordnet} taxonomy, and also retrieve the sub-tree from WordNet that perfectly contains our concepts and use this as our ground-truth taxonomy on the basis of which we carry out subsequent experiments.
We generate our negative samples by choosing a measure derived primarily from the Wu-Palmer similarity \citep{wu-palmer-1994-verb}.
This similarity can be computed over any taxonomy using the following operations:
% \todo{We use similarities across the is-a hierarchy to obtain negative samples for each property}. 
\begin{align}
    sim_{\texttt{wup}}(c_i, c_j) = \frac{2 \times \depth(\lcs(c_i, c_j))}{\depth(c_i) + \depth(c_j)},
\end{align}
where $\lcs(x_1, x_2)$ is a function that computes the least-common subsumer\footnote{a node in the hierarchy that is a hypernym/parent of the input concepts with minimum depth. For instance, $\lcs(\textsc{robin}, \textsc{bat}) = \textsc{vertebrate}$.} of the two\footnote{although in practice it can be applied for multiple concepts.} concepts, and $\depth(x)$ computes the length of the path between the input concept and the root node of the hierarchy. 
We consider a generalized form of this measure (denoted as $sim_{\texttt{gwup}}$), to compute the similarity of a single concept to a set of concepts:
\begin{align}
    % sim_{\texttt{gwup}}(c_1, \dots, c_n, r) = \frac{(n + 1) \times \depth(\lcs(c_1, \dots, c_n, r))}{\depth(c_1) + \dots + \depth(c_n) + \depth(r)}
    sim_{\texttt{gwup}}(c_1, \dots, c_n) &= \frac{n \times \depth(\lcs(c_1, \dots, c_n))}{\depth(c_1) + \dots + \depth(c_n)}
    % sim_{\texttt{gwup}}([c_1, \dots, c_n], r_i) &= \frac{n \times \depth(\lcs(c_1, \dots, c_n))}{\depth(c_1) + \dots + \depth(c_n)}
\end{align}
For every property $P_i$, we use this measure in \cref{alg:propjudgdataset} to sample $k$ concepts from $\leftovers_{P_i}$, based on their $sim_{gwup}$ with $\positives_{P_i} = \{c_1, \dots, c_k\}$.
For example, consider the property \textit{has striped patterns on its body}, the corresponding artifacts would be:
\begin{align*}
    \positives &= \{\textsc{zebra}, \textsc{tiger}, \textsc{bee}, \textsc{wasp}\}\\
    \leftovers &= \concepts - \positives\\
    &= \{\textsc{accordion}, \dots, \textsc{yo-yo}\}\\
    \mathrm{NS} = \negsamp(\leftovers, 4) &= \{\textsc{horse}, \textsc{lion}, \textsc{ant}, \textsc{beetle}\}
\end{align*}
\begin{align*}
    \mathcal{D} = \{&[\textit{a zebra has striped patterns on its body}, \true],\\&\dots, \\
    &[\textit{a beetle has striped patterns on its body}, \false]\}
\end{align*}
Note that we follow the method outlined by \citet{bhatia2020transformer} to convert concept-property pairs into sentences, which we denote as $\textit{sentencizer}()$ in \Cref{alg:propjudgdataset}.

\begin{algorithm}[H]
\textbf{Input:} $\concepts = \{c_1, \dots, c_n\}$: Set of all concepts, $n = 521$.\\
\hspace*{2.9em}$\properties = \{P_1, \dots, P_m\}$: Set of all properties, $m = 3735$.
\begin{algorithmic}[1]
\State $\mathcal{D} \gets [\,]$ \rightcomment{the final set of stimuli for the property judgment task.}
% \LineComment{hmm.}
% \State $2$  \rightcomment{Initialize output value to $1$}
\For{$i = 1, \ldots, m$}
    \State $\positives_{P_i} \gets [c_1, \dots, c_k]$ \rightcomment{set of $k$ concepts that possess the property $P_i$}
    \State $\leftovers_{P_i} \gets \concepts - \positives_{P_i}$
    \LineComment{Lines 6--9 compute $\negsamp(\leftovers_{P_i}, k)$}
    \State $\mathrm{NS}_{P_i} \gets [\,]$ \rightcomment{set of negative samples for the property $P_i$}
    \State $\Tilde{\leftovers_{P_i}} \gets \text{argsort}(\leftovers_{P_i}, sim_{\texttt{gwup}})$ \rightcomment{sort $\leftovers_{P_i}$ based on $sim_{\texttt{gwup}}(c_1, \dots, c_k, x_j)\, \forall x_j \in \leftovers_{P_i}$}
    \For{$j = 1, \ldots, k$}
        \State $\mathrm{NS}_{P_i}.\text{append}(\Tilde{\leftovers_{P_i}}[j])$ \rightcomment{take the top $k$ concepts from $\leftovers_{P_i}$ as negative samples}
    \EndFor
% \For{$k = K, \ldots, 1$}
%     \State $3$
%     \State $2$
%     \State $1$
% \EndFor
\LineComment{the following pairs the positive and negative samples with their labels, and appends them to $\mathcal{D}$}
    \For{$j = 1, \ldots, k$}
    \LineComment{$\textit{sentencize}()$ constructs a sentence using a concept and a property-phrase \citep[see][]{bhatia2020transformer}.}
        \State $\mathcal{D}.\text{append}([\textit{sentencize}(\positives_{P_i}[j], P_i), \true])$
        \State $\mathcal{D}.\text{append}([\textit{sentencize}(\mathrm{NS}_{P_i}[j], P_i), \false])$
    \EndFor
    % $\mathcal{D}.\text{extend}(x)$
\EndFor
% \LinesComment{Apply \cref{eq:incl-from-grad}}
% \State $\pi \gets \boldsymbol{0}^N$
% \For{$n = 1 \ldots N$}
% \State $\pi \gets \frac{w_n}{Z}$
% \EndFor
\State \Return $\mathcal{D}$
\end{algorithmic}
\caption{Algorithm to generate the dataset, $\mathcal{D}$, for the property judgment task}
\label{alg:propjudgdataset}
\end{algorithm}

\section{Linear Mixed Effects Model Results}
We use a linear-mixed effects models to test the connection between LMs’ generalization behavior and the overlap in training data properties
For each model, we use the following LMER specification \citep{lme4}:
\begin{quote}
    \centering
    \texttt{G $\sim$ n + overlap * sim + (1|property) + (1|trial)},
\end{quote}
where:
\begin{itemize}
    \item \texttt{G} is the generalization score (see Eq. 1 in the paper).
    \item \texttt{n} is the number of adaptation concepts (i.e., the number of premise statements).
    \item \texttt{overlap} is the property overlap between the adaptation and the generalization set in each trial, calculated as the jaccard similarity between the binary property-vectors of each concept.
    \item \texttt{sim} is the cosine similarity between the embeddings (from the pre-contextualized layer in each model) of the concepts in the adaptation and generalization sets in each trial. Note that this is a model-dependent measure.
    \item \texttt{property} is the novel property (one out of 8) that is projected in the trial.
    \item \texttt{trial} is the individual trial.
\end{itemize}
In what follows, we report results from fitting this model to the results and statistics of our three property-induction models. We use Satterthwaite's method \citep{lmertest} to perform significance testing.
% \subsection{ALBERT-xxlarge}

\begin{table}[!h]
\centering
\begin{tabular}{@{}lcccccc@{}}
\toprule
Fixed-effect           & $\beta$   & $SE$     & $t$     & $df$      & $p$      \\ \midrule
\texttt{n}             & $0.0544$  & $0.0035$ & $15.41$ & $304.59$  & $<2e-16$ \\
\texttt{overlap}       & $0.3951$  & $0.0229$ & $17.26$ & $7123.98$ & $<2e-16$ \\
\texttt{sim}           & $0.1102$  & $0.0180$ & $6.11$  & $6751.08$ & $1e-9$ \\
\texttt{overlap * sim} & $0.8583$ & $0.2263$ & $3.79$ & $7170.65$ & $0.0001$  \\ \bottomrule
\end{tabular}
\caption{Results for ALBERT-xxl}
\label{tab:albert}
\end{table}

% \subsection{BERT-large}

\begin{table}[!h]
\centering
\begin{tabular}{@{}lcccccc@{}}
\toprule
Fixed-effect           & $\beta$   & $SE$     & $t$     & $df$      & $p$      \\ \midrule
\texttt{n}             & $0.0589$  & $0.0046$ & $12.76$ & $396.84$  & $<2e-16$ \\
\texttt{overlap}       & $0.4731$  & $0.0245$ & $19.28$ & $7088.53$ & $<2e-16$ \\
\texttt{sim}           & $0.1696$  & $0.0487$ & $3.48$  & $6509.68$ & $0.0005$ \\
\texttt{overlap * sim} & $0.7429$ & $0.3415$ & $2.18$ & $7180.52$ & $0.03$  \\ \bottomrule
\end{tabular}
\caption{Results for BERT-large}
\label{tab:bert}
\end{table}

% \subsection{RoBERTa-large}

\begin{table}[!h]
\centering
\begin{tabular}{@{}lcccccc@{}}
\toprule
Fixed-effect           & $\beta$   & $SE$     & $t$     & $df$      & $p$      \\ \midrule
\texttt{n}             & $0.0555$  & $0.0051$ & $10.79$ & $404.43$  & $<2e-16$ \\
\texttt{overlap}       & $0.3851$  & $0.0269$ & $14.32$ & $7187.79$ & $<2e-16$ \\
\texttt{sim}           & $0.3631$  & $0.0665$ & $5.46$  & $6180.44$ & $4.9e-8$ \\
\texttt{overlap * sim} & $-1.1735$ & $0.4164$ & $-2.82$ & $7186.19$ & $0.005$  \\ \bottomrule
\end{tabular}
\caption{Results for RoBERTa-large}
\label{tab:roberta}
\end{table}

% \section{Additional Observations}
% \paragraph{Embedding space similarities of the models do not track category membership} We find
% % figure showing outside to be greater similarity than taxonomy -- existence proof that cosine similarity does not track taxonomic membership.
% \paragraph{Relation to Osherson et al.'s SimCov Model}
\clearpage
\bibliographystyle{apacite}

\bibliography{CogSci_Template}